\documentclass[11pt]{article}

\usepackage[preprint]{acl}

\usepackage{times}
\usepackage{latexsym}

\usepackage[T1]{fontenc}

\usepackage[utf8]{inputenc}

\usepackage{microtype}

\usepackage{inconsolata}

\usepackage{graphicx}
\graphicspath{{figures/}}

\usepackage{amsmath}
\usepackage{amssymb}

\usepackage{booktabs}
\usepackage{multirow}
\usepackage{placeins}
\usepackage{xcolor}

\newcommand{\methodname}{JEDI}
\newcommand{\cflow}{ContextFlow}
\newcommand{\pizero}{\ensuremath{\pi_0}}
\newcommand{\vjepa}{V-JEPA~2.1}
\newcommand{\liberoorder}{LIBERO-Order}

\newcommand{\support}{\mathcal{C}}            %
\newcommand{\comp}{E_\phi}                     %
\newcommand{\pred}{g_\psi}                     %
\newcommand{\bank}{\mathcal{B}}               %
\newcommand{\tgtenc}{\bar f}                   %
\newcommand{\slots}{Z_{\support}}              %
\newcommand{\Ljepa}{\mathcal{L}_{\text{JEPA}}}
\newcommand{\Lcf}{\mathcal{L}_{\text{cf}}}
\newcommand{\Lcfm}{\mathcal{L}_{\text{CFM}}}
\newcommand{\DDtask}{\text{DD}_{\text{task}}}
\newcommand{\DDproc}{\text{DD}_{\text{proc}}}

\newcommand{\LN}{\operatorname{LN}}

\newcommand{\best}[1]{{\bfseries\boldmath #1}}
\newcommand{\na}{--}

\newcommand{\pstd}[1]{\ensuremath{\pm #1}}

\newcommand{\draftvalue}[1]{\textcolor{black}{#1}}

\makeatletter
\@ifpackageloaded{lineno}{%
  \IfFormatAtLeastTF{2026-06-01}{}{%
    \IfFormatAtLeastTF{2025-06-01}{\RemoveFromHook{build/column/before}[lineno]}{}%
    \AtBeginDocument{%
      \def\@LN@makecol{%
        \@LN@orig@makecol
        \setbox\@outputbox\vbox{%
          \boxmaxdepth\@maxdepth
          \protected@write\@auxout{}{%
            \string\def\string\@LN@column{\if@firstcolumn1\else2\fi}}%
          \box\@outputbox}}}}}{}
\makeatother

\title{\methodname{}: Joint-Embedding Demonstration Compression\\for In-Context Imitation Learning}

\author{Jiazhao Liang$^{1}$\thanks{Equal contribution.} \quad Zhewen He$^{2}$\footnotemark[1] \quad Shuaihang Yuan$^{2}$ \quad Hao Huang$^{2}$ \quad Yi Fang$^{2}$ \\ $^{1}$New York University Tandon School of Engineering, Brooklyn, NY 11201, USA \\ $^{2}$NYUAD Center for Artificial Intelligence and Robotics (CAIR) and Embodied AI and Robotics (AIR) Lab, \\ New York University Abu Dhabi, Abu Dhabi 129188, UAE}

\makeatletter
\AtBeginDocument{%
  \g@addto@macro\@maketitle{%
    \par\begin{minipage}{\textwidth}
  \centering
  \captionsetup{hypcap=false}
  \includegraphics[width=0.9\linewidth]{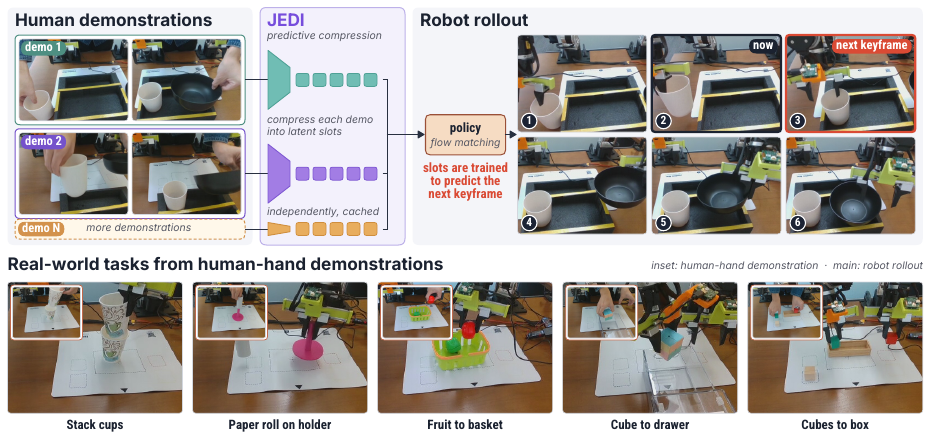}
  \captionof{figure}{\textbf{Many demonstrations, compressed by prediction.} \methodname{} compresses each demonstration independently into $k$ latent slots, which are cached and trained to predict the next keyframe of the query episode; a flow-matching policy attends to the slots and acts on the robot.
  \textbf{Top:} human demonstrations of one task, their compression, and keyframes of the resulting robot rollout.
  \textbf{Bottom:} five real-world tasks executed from human-hand demonstrations (inset: demonstration; main: robot rollout).}
  \label{fig:teaser}
\end{minipage}
\par\vspace{\baselineskip}}}
\makeatother

\begin{document}
\maketitle

\begin{abstract}
In-context imitation learning (ICIL) adapts robot policies through test-time demonstrations, but long multimodal trajectories constrain how many examples fit in context. We ask what a compressed demonstration should preserve. We propose \methodname{}, which compresses demonstrations into latent slots trained to predict query episode's next-stage keyframe in a frozen video-JEPA space. Task and process counterfactuals help the slots preserve execution details, while the predictor selects demonstrations from a bank. Post-training with keypoint tokens enables human-hand conditioning on robots. Experiments on LIBERO and real robots show that \methodname{} supports many-shot context; LIBERO objective controls show improved success and demonstration following over action-only compression. These findings support query-future prediction as a training objective for preserving execution-relevant information in compressed demonstrations.
\end{abstract}

\section{Introduction}
\label{sec:intro}

In-context learning lets a language model adapt to a new task from examples in its prompt, without weight updates~\citep{brown-etal-2020-gpt3}.
In-context imitation learning (ICIL) extends this ability to robot manipulation: at test time a policy receives demonstrations of an unseen task configuration and executes it without fine-tuning~\citep{fu-etal-2025-icrt,vosylius-johns-2025-instant,ding-etal-2026-contextflow}.
Its scope is unseen objects, arrangements and goals composed of motion primitives seen in training.

The bottleneck is the context itself.
A demonstration is a multi-view sensorimotor trajectory spanning hundreds of control steps and still takes thousands of tokens after subsampling, so current ICIL policies typically use only one or two demonstrations, while transformer policies degrade when prompts exceed the lengths seen during training~\citep{yoo-etal-2025-robossm}.
Compression alleviates this cost: \cflow{} compresses each demo with perceiver-style compressors~\citep{ding-etal-2026-contextflow}, but its compressor is trained only through the action loss and is evaluated with a single demonstration.

This leaves open \emph{what a demonstration compressor should retain}.
Pooled task embeddings~\citep{james-etal-2018-tecnets,jang-etal-2022-bcz} are inexpensive but retain little beyond the intent of the task, whereas the full sensorimotor context retains the stage-dependent detail that control needs, at a cost linear in the number of demonstrations~\citep{li-2026-survey}.
Language models face the same question: prompts are compressed into gist tokens~\citep{mu-etal-2023-gist} or by autoencoding~\citep{ge-etal-2024-icae}, and in the many-shot regime such compressors can underperform the simple baseline of using fewer shots~\citep{khatri-etal-2025-memcom}.
A compressor's training objective, not only its architecture, decides what is kept.

We argue that a compressor should retain the information that allows the policy to \emph{anticipate the next stage of its own episode}, and propose \methodname{} (\textbf{J}oint-\textbf{E}mbedding \textbf{D}emonstration compression for \textbf{I}n-context imitation learning).
\methodname{} compresses each demonstration into $k$ latent slots, from which a predictor forecasts the query episode's next-stage keyframe in the embedding space of a frozen video JEPA, \vjepa{}~\citep{assran-etal-2025-vjepa2,murlabadia-etal-2026-vjepa21}.
Because most of a future frame is predictable from the current one, a counterfactual term requires the prediction to be worse when the slots come from a demonstration of another task, or of the same task performed differently.
The slots therefore carry the \emph{process} of the demonstration, such as object order and grasp height, and not only its task identity.
Each demonstration is compressed independently, so its slots can be cached and reused, and a policy can attend to many demonstrations (Figure~\ref{fig:teaser}).

At test time the query's future is unknown, but the bank demonstrations' futures are known.
We score each demonstration by how well its slots predict the others' keyframes, extending predictive selection of language-model in-context examples to embodied demonstrations~\citep{rubin-etal-2022-learning}.

Success rate alone does not reveal what a policy uses from its context.
We adapt the analysis of in-context learning in language models, which separates recognizing a task from learning it~\citep{min-etal-2022-rethinking,pan-etal-2023-context}, and the paired-context protocol recommended for ICIL~\citep{li-2026-survey}: on the same checkpoints we swap demonstrated goals and, separately, vary the process at a fixed endpoint.
Across these tests, predictive compression keeps stage-dependent detail that action-only compression loses and follows swapped demonstrations 40.8 points more often (\S\ref{sec:analysis}).

Our contributions are as follows.
(1)~\textbf{Predictive compression.} A per-demonstration compressor is trained with a demonstration-conditioned predictive objective in a frozen video-JEPA space, together with task and process counterfactuals (\S\ref{sec:method}). After policy training, these predictively pretrained, frozen slots achieve 75.4\% success (\S\ref{sec:objective-ablation}).
(2)~\textbf{Predictive demonstration selection.} The predictor decides which demonstrations to place in context based on its prediction error. To separate the effect of \emph{which} demonstrations are chosen from \emph{how many} are used, we compare all methods at the same number of demonstrations ($N$). Across three seeds, this selection beats random choice and, when the demonstration pool is diverse, also beats image-based retrieval (\S\ref{sec:selection}, \S\ref{sec:manyshot}).
(3)~\textbf{Post-training with human-hand demonstrations.} Keypoint-based post-training supports human-hand context on a dual-arm platform, without test-time weight updates (\S\ref{sec:posttrain}, \S\ref{sec:real}).
(4)~\textbf{Language and demonstrations.} \methodname{} follows the demonstrations where the instruction is silent and the instruction where it is explicit, more reliably than action-only compression on \liberoorder{} (\S\ref{sec:language}).

\section{Related Work}
\label{sec:related}

\paragraph{In-context imitation learning.}
Demonstration-conditioned policies date back to one-shot imitation~\citep{duan-etal-2017-oneshot,dasari-gupta-2021-oneshot} and task-embedding networks~\citep{james-etal-2018-tecnets,jang-etal-2022-bcz}.
Recent ICIL methods treat demonstrations as a prefix for next-token prediction~\citep{fu-etal-2025-icrt,vuong-etal-2025-tokenizer,nguyen-etal-2026-iclr}, expose spatial structure through graphs, keypoints, object poses or dense semantic matching~\citep{vosylius-johns-2025-instant,dipalo-johns-2024-kat,yin-etal-2025-roboprompt,she-etal-2026-matchingpolicy}, retrieve stored trajectories~\citep{sridhar-etal-2025-regent,sridhar-etal-2025-ricl,sato-etal-2026-sail}, or use a state-space model so that long prompts remain usable~\citep{yoo-etal-2025-robossm}.
Perceiver and Q-Former bottlenecks~\citep{jaegle-etal-2021-perceiver,jaegle-etal-2022-perceiverio,li-etal-2023-blip2} shorten context: \cflow{} uses perceiver compressors with flow matching~\citep{ding-etal-2026-contextflow}, and Behavior Prompting Policy attention-pools chunk embeddings~\citep{patel-etal-2026-bpp}.
Both train their compressors through the action loss and use one demonstration.

\paragraph{Predictive representations and human videos.}
JEPAs predict target representations instead of reconstructing inputs~\citep{assran-etal-2023-ijepa,bardes-etal-2024-vjepa,assran-etal-2025-vjepa2,murlabadia-etal-2026-vjepa21}, an idea since extended to language-model training~\citep{huang-etal-2025-llmjepa}.
Prediction in pretrained feature spaces supports robot world models and auxiliary policy targets~\citep{zhou-etal-2025-dinowm,lin-etal-2026-jepawam,vujinovic-kovacevic-2026-actjepa}.
FLARE aligns action-denoiser features with future observation embeddings~\citep{zheng-etal-2025-flare}.
Demo-JEPA predicts target-embodiment latents from one cross-embodiment demonstration for planning, without compressing the demonstration~\citep{he-etal-2026-demojepa}.
SeeTraceAct trains an observation-conditioned visual latent plan with action prediction and visibility-aware future end-effector traces~\citep{son-etal-2026-seetraceact}.
ViVLA jointly predicts demonstration and agent latent actions in a shared discrete space, alongside robot actions~\citep{chen-etal-2025-vivla}.
\methodname{} independently compresses demonstrations into cached slots trained to predict the query's next keyframe in frozen video-JEPA space with task/process counterfactuals; the predictor also selects demonstrations.
Human videos are a second source of demonstrations: a policy can be conditioned on one such video through meta-learning or a video encoder~\citep{yu-etal-2018-daml,jang-etal-2022-bcz,jain-etal-2024-vid2robot} or co-trained on egocentric human data~\citep{kareer-etal-2024-egomimic}, and keypoints or motion tracks give hand and gripper a shared representation~\citep{papagiannis-etal-2025-rx,ren-etal-2025-motiontracks}; MimicDroid learns in context from human play videos~\citep{shah-etal-2025-mimicdroid}.
We admit human-hand videos to a compressed context by post-training with keypoint tokens and the predictive objective (\S\ref{sec:posttrain}).

\paragraph{In-context learning in language models.}
Prompts are compressed into gist tokens~\citep{mu-etal-2023-gist}, by autoencoding~\citep{ge-etal-2024-icae} or into summary vectors~\citep{chevalier-etal-2023-adapting}; in the many-shot regime~\citep{agarwal-etal-2024-manyshot}, however, existing prompt compressors are ineffective and using fewer shots is a strong baseline~\citep{khatri-etal-2025-memcom}.
In-context examples are selected by similarity to the query~\citep{liu-etal-2022-makes} or by how well the model predicts a held-out example when conditioned on them~\citep{rubin-etal-2022-learning}.
Closest in design, UniICL compresses each demonstration independently into memory tokens, caches them in a demonstration bank and selects by their similarity to the query~\citep{gao-etal-2024-uniicl}; we differ in the objective that trains the slots and in selecting by prediction error.
Analyses of what a model uses from its context showed that the correctness of in-context labels matters less than the format and the input distribution~\citep{min-etal-2022-rethinking}, separated task recognition from task learning~\citep{pan-etal-2023-context}, and summarized a context as task or function vectors~\citep{hendel-etal-2023-context,todd-etal-2024-function}.
Comparable analyses for robots are scarce and mostly retrain the model per condition; \citet{li-2026-survey} recommends paired contexts scored on both targets, same-endpoint/different-process tests, and variance over support draws, which we run on the same checkpoints across compression objectives.

\section{Method}
\label{sec:method}

\begin{figure*}[t]
  \centering
  \includegraphics[width=0.9\textwidth]{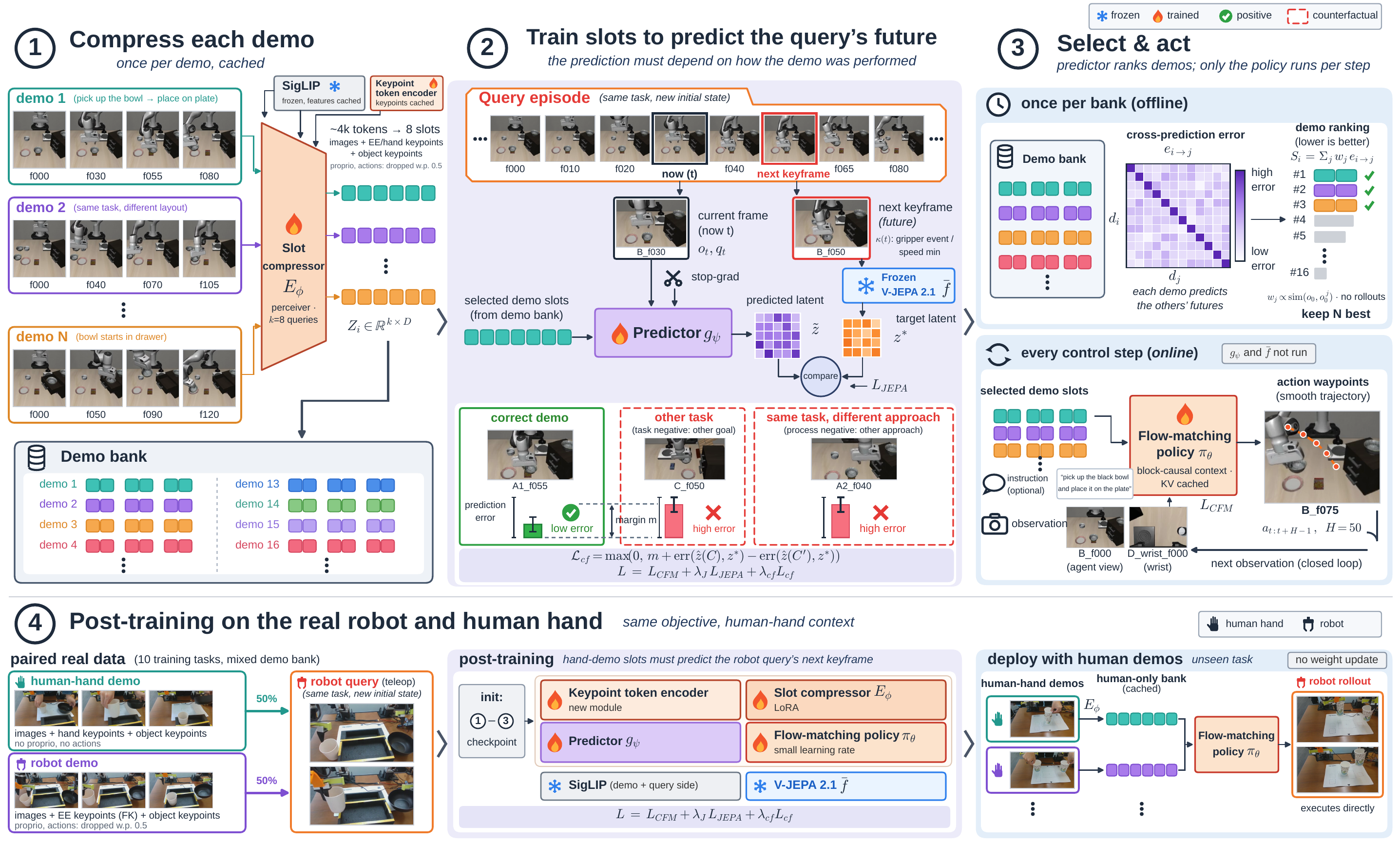}
  \caption{\textbf{Overview of \methodname{}.}
  \textbf{(1)} A compressor maps each demonstration to $k$ latent slots, which are cached in a demonstration bank.
  \textbf{(2)} The slots are trained to predict the next keyframe of the query episode in the embedding space of a frozen \vjepa{} encoder, with task and process negatives.
  \textbf{(3)} The predictor ranks the demonstrations in the bank, and only the policy runs in the control loop.
  \textbf{(4)} Post-training on 10 tasks uses equal shares of human-hand and dual-arm robot demonstrations with the same objective; unseen tasks are executed from human-hand context without test-time weight updates.}
  \label{fig:method}
\end{figure*}

\subsection{Problem Setup}
\label{sec:problem}

A \emph{task configuration} $\tau$ is a goal together with an object set and a scene.
A \emph{demonstration} is a trajectory $d=\{(I_t,q_t,a_t)\}_{t=1}^{T}$ of multi-view images, proprioception and actions; on the real robot, $d$ may also be a human-hand video without $q_t$ and $a_t$ (\S\ref{sec:posttrain}).
Given a \emph{support set} $\support=\{d_1,\dots,d_N\}$ of $\tau$ and an optional instruction $\ell$, the policy $\pi_\theta(a_{t:t+H-1}\mid I_t,q_t,\slots,\ell)$ acts in a \emph{query episode}, a fresh rollout of $\tau$ from a new initial state; $\slots$ is the context built from $\support$, and $\theta$ is fixed at test time.
We evaluate on \emph{unseen configurations}, which are held out from training but share motion primitives with it.
The support set is drawn from a \emph{demonstration bank} $\bank=\{d_1,\dots,d_M\}$, $M\ge N$, which may also contain demonstrations of other configurations.
We call $N\ge 4$ the many-shot regime.
A \emph{process} is the manner in which a task is executed, such as the order in which objects are handled, the grasp location and height, and the approach direction; a faithful policy should follow it, not only reach the goal (notation in Appendix~\ref{app:notation}).

\methodname{} has three trained components (Figure~\ref{fig:method}): a per-demonstration compressor $\comp$, a predictor $\pred$ and a flow-matching policy $\pi_\theta$; a frozen video encoder $\tgtenc$ supplies the prediction target.
The predictor defines the training objective (\S\ref{sec:objective}--\ref{sec:cf}) and selects demonstrations at test time (\S\ref{sec:selection}).

\subsection{Per-Demonstration Compressor}
\label{sec:compressor}

Following \cflow{}, each demonstration is subsampled to $T'{=}8$ image frames and $128$ proprioception and action steps, which are encoded by SigLIP~\citep{zhai-etal-2023-siglip}, the vision encoder of the policy, and by MLPs (Appendix~\ref{app:impl}).
The compressor is a perceiver~\citep{jaegle-etal-2021-perceiver} with $k$ learnable queries and $L_c{=}4$ layers, and outputs slots $Z_i\in\mathbb{R}^{k\times D}$ of the policy's token width $D$.
The context is the concatenation $\slots=[Z_1;\dots;Z_N]$; slots carry a learned slot-index embedding but no demonstration index, so reordering demonstrations only reorders their slot blocks.
Because $\comp$ sees one demonstration at a time, slots are computed once per demonstration and cached.

\subsection{Demonstration-Conditioned Predictive Objective}
\label{sec:objective}

\paragraph{Target.}
We segment every episode into stages with a keyframe heuristic adapted from prior work~\citep{shridhar-etal-2022-peract,yin-etal-2025-roboprompt}: \emph{keyframes} are gripper open/close events, near-stationary steps (local minima of end-effector speed) and the end of the episode.
Let $\kappa(t)$ be the first keyframe after a non-terminal query step $t$.
The target is the latent of a two-frame clip at that keyframe under a frozen \vjepa{} encoder, $P$ patch tokens of width $D_z$:
\begin{equation}
  z^\star=\tgtenc\big(I^{\text{query}}_{\kappa(t)-1:\kappa(t)}\big)\in\mathbb{R}^{P\times D_z}.
\end{equation}
Targets are computed once offline with frozen $\tgtenc$ and stay fixed during training.

\paragraph{Predictor and loss.}
A small transformer $\pred$ takes the slots $\slots$, the current query observation tokens $o_t$ from the policy's SigLIP encoder and the proprioception $q_t$, and outputs $\hat z$ with the patch layout of $z^\star$.
It does not receive the offset $\Delta=\kappa(t)-t$, which is derived from the query's own future and would lower the loss without any use of the demonstrations.
With layer normalization $\LN$ and the SmoothL1 penalty $h$ summed over features,
\begin{equation}
  \Ljepa=\frac{1}{P}\sum_{p=1}^{P}h\big(\LN(\hat z_p)-\LN(z^\star_p)\big).
  \label{eq:jepa}
\end{equation}
A deterministic regression suffices because the next keyframe is nearly deterministic given the demonstrated process and the current state, which avoids the failure of deterministic latent prediction on multimodal targets~\citep{dinh-vo-2026-jepaparadox}.
Gradients flow into $\pred$ and $\comp$ only: $o_t$ enters with a stop-gradient, so these losses do not train the policy's observation encoder (hyperparameters in Table~\ref{tab:hparams}).

\subsection{Counterfactual Demonstration Term}
\label{sec:cf}

Most of $z^\star$ is predictable from the current frame alone, so the predictor can lower $\Ljepa$ while ignoring $\slots$.
We therefore require the prediction to degrade when the support set is replaced by a counterfactual one, $\support'$, of two types.
A \emph{task negative} contains demonstrations of a different task in the same scene, and forces the slots to encode the target object and the goal.
A \emph{process negative} contains demonstrations of the same task performed with a different process, chosen so that the query's next keyframe differs; it forces the slots to encode order, grasp and approach.
With $\operatorname{err}(\hat z,z^\star)$ the value of Eq.~\eqref{eq:jepa} for one sample, $\operatorname{err}(\support)$ its value for a prediction from $\support$, $\mathcal{N}$ the negatives and $m$ a margin,
\begin{equation}
  \Lcf=\frac{1}{|\mathcal{N}|}\sum_{\support'\in\mathcal{N}}\max\!\big(0,\ m+\operatorname{err}(\support)-\operatorname{err}(\support')\big).
  \label{eq:cf}
\end{equation}
Process labels come from construction where a benchmark provides them and from clustering each task's keyframe-latent sequences otherwise (Appendix~\ref{app:keyframes}); they are used only to choose negatives, and positive support sets always share the query's process.
As a diagnostic we report the \emph{demonstration dependence} $\text{DD}=\operatorname{err}(\support')/\operatorname{err}(\support)$ separately for task negatives ($\DDtask$) and process negatives ($\DDproc$); a value near~1 means it is ignored.

\subsection{Policy and Training}
\label{sec:policy}

The policy follows \pizero{}~\citep{black-etal-2025-pi0} and \cflow{}: a flow-matching action expert~\citep{lipman-etal-2023-flow} with action horizon $H{=}50$ and 10 Euler steps, initialized from \pizero{} and adapted with LoRA~\citep{hu-etal-2022-lora}.
It attends to $[\slots;o_t;\ell]$ through a context expert with block-causal attention, so that the keys and values of the context are cached across flow steps.
With noise $\epsilon$, flow time $s$ and $a^s_t=s\,a_t+(1-s)\,\epsilon$,
\begin{equation}
  \Lcfm=\mathbb{E}\,\big\|u_\theta(a^s_t,o_t,q_t,\slots,\ell,s)-(a_t-\epsilon)\big\|^2 .
\end{equation}
The total loss is $\mathcal{L}=\Lcfm+\lambda_J\,\Ljepa+\lambda_{\text{cf}}\,\Lcf$.
Support and query are disjoint episodes of the same configuration, and of the same process when process labels exist.
We sample $N\sim\mathcal{U}\{1,\dots,16\}$, since training at a fixed prompt length fails to extrapolate~\citep{yoo-etal-2025-robossm}.
With probability $0.2$ we add one to four distractor demonstrations from other configurations, and with probability $0.5$ we replace the instruction by an empty string, so that demonstration-only conditioning lies inside the training distribution.
SigLIP is frozen on the demonstration side, so its features can be cached, and fine-tuned on the query side.

\subsection{Predictive Demonstration Selection}
\label{sec:selection}

\paragraph{Bank consistency.}
At test time the bank can hold more demonstrations than the context and contain demonstrations that are irrelevant or performed differently.
For sensorimotor banks, we choose the $N$ demonstrations with the predictor, as language-model examples are scored by how well they predict a held-out example~\citep{rubin-etal-2022-learning}.
The future of the query is unknown, but the futures of the demonstrations in the bank are known, so we treat each demonstration $d_j$ as a pseudo-query and measure how well the slots of another demonstration $d_i$ predict its keyframes,
\begin{equation}
  e_{i\to j}=\frac{1}{|\mathcal{T}_j|}\sum_{t\in\mathcal{T}_j}\operatorname{err}\big(\pred(Z_i,o^{j}_t,q^{j}_t),\,z^\star_{j,\kappa(t)}\big),
  \label{eq:consistency}
\end{equation}
where $\mathcal{T}_j$ are the subsampled steps of $d_j$.
This is the training loss of \S\ref{sec:objective} evaluated inside the bank, on disjoint episodes as in training; it needs no rollouts and no information about the query, so the matrix $(e_{i\to j})$ is computed once per bank and cached.

\paragraph{Selection score.}
We weight the other demonstrations by the similarity of their initial observations $o^{j}_0$ to the query's initial observation $o_0$,
\begin{equation}
  S_i=\sum_{j\ne i}w_{ij}\,e_{i\to j},\quad w_{ij}\propto e^{\,\text{sim}(o_0,o^{j}_0)/\eta},
  \label{eq:score}
\end{equation}
where sim is the cosine similarity of pooled image features and $\eta$ a temperature. For each $i$, $\sum_{j\ne i}w_{ij}=1$; we select the $N$ lowest scores.
Uniform weights measure agreement with the rest of the bank and filter out distractors; peaked weights favour demonstrations that are predictive in layouts similar to that of the query.
The score assumes that most of the bank shows the intended configuration.
When a bank contains several processes for one task, we add each completed stage of the query as a further pseudo-query and reselect (\emph{online commitment}, Appendix~\ref{app:impl}), which keeps the context on the process that the policy has begun.
Slots and the consistency matrix are cached; $\pred$ and $\tgtenc$ run once per bank and, with online commitment, once per stage, never in the control loop.

\subsection{Post-Training with Human-Hand Demonstrations}
\label{sec:posttrain}

On the real robot a demonstration may be a human-hand video, which shows another embodiment and has neither proprioception nor actions.
We post-train the checkpoint of the preceding stages on real data (Figure~\ref{fig:method}, stage~4) with the same loss $\mathcal{L}$.
A keypoint token encoder adds to the input of the compressor keypoints that exist for both embodiments: the wrist, the thumb tip, the index fingertip and the pinch aperture of the hand, the corresponding end-effector keypoints of the robot from forward kinematics, and object keypoints from one offline tracker; robot proprioception and action tokens are dropped with probability $0.5$.

Each training bank mixes human-hand and robot demonstrations in equal shares, while the query is always a robot episode of the same task.
The slots of a human-hand demonstration are therefore trained to predict the next keyframe of the robot: the predictive objective asks what a demonstration implies for the future of the query, not how it looks, so it also relates the two embodiments.
SigLIP and $\tgtenc$ stay frozen; the keypoint token encoder and the predictor are trained, the compressor is updated with LoRA and the policy with a reduced learning rate.
At deployment the bank may hold human-hand demonstrations only, with no weight updates (Appendix~\ref{app:real}); simulation omits this stage.

\section{Experiments}
\label{sec:experiments}

\begin{table*}[t]
  \centering
  \small
  \setlength{\tabcolsep}{5pt}
  \begin{tabular}{@{}lcccc@{}}
    \toprule
    \multirow{2}{*}{Method} & \multicolumn{3}{c}{LIBERO (\%, $\uparrow$)} & \multirow{2}{*}{RLBench (\%, $\uparrow$)} \\
    \cmidrule(lr){2-4}
    & Spatial & Object & Goal & \\
    \midrule
    ICRT~\citep{fu-etal-2025-icrt} & 58.5\pstd{3.4} & 44.6\pstd{4.6} & 24.6\pstd{2.1} & 49.3\pstd{3.8} \\
    RoboSSM~\citep{yoo-etal-2025-robossm} & 66.8\pstd{3.1} & 54.3\pstd{4.3} & 37.3\pstd{3.6} & 54.4\pstd{2.8} \\
    Instant Policy~\citep{vosylius-johns-2025-instant} & 83.1\pstd{3.2} & 69.1\pstd{2.9} & 47.1\pstd{2.8} & 77.1\pstd{3.3} \\
    \cflow{}~\citep{ding-etal-2026-contextflow} & 85.6\pstd{2.8} & 76.4\pstd{3.2} & 56.5\pstd{2.2} & 73.6\pstd{1.3} \\
    \methodname{} (ours) & \best{92.7\pstd{1.6}} & \best{83.6\pstd{2.9}} & \best{63.0\pstd{2.3}} & \best{82.1\pstd{2.0}} \\
    \midrule
    \methodname{}, $N{=}16$ & 95.4\pstd{\draftvalue{1.0}} & 86.2\pstd{\draftvalue{1.1}} & 69.1\pstd{\draftvalue{2.5}} & 86.3\pstd{\draftvalue{1.7}} \\
    \bottomrule
  \end{tabular}
  \caption{\textbf{Success rate on unseen manipulation configurations} in LIBERO-Spatial, LIBERO-Object, LIBERO-Goal and RLBench.
  The first five rows use one demonstration ($N{=}1$); input adapters are distinguished in Appendix~\ref{app:baselines}.
  Values are mean $\pm$ standard deviation (\%) over 3 training seeds.
  Each LIBERO seed uses 4 held-out tasks and 100 rollouts per task (1,200 trials per suite across seeds; Appendix~\ref{app:eval}).
  The best $N{=}1$ mean in each column is in bold.}
  \label{tab:main}
  \label{tab:n16}
\end{table*}

\begin{figure*}[t]
  \centering
  \begin{minipage}[t]{0.585\textwidth}
  \vspace{0pt}\centering
  \includegraphics[width=\linewidth]{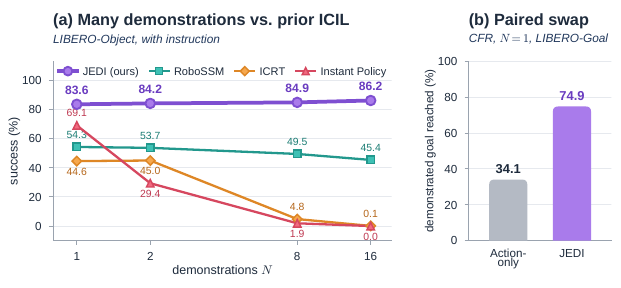}
  \caption{\textbf{Many-shot performance and context use.}
  (a) LIBERO-Object success vs.\ $N$; ICRT and Instant Policy are designed for one or few demonstrations; points beyond $N{=}2$ exceed their intended range.
  (b) CFR under paired swaps on LIBERO-Goal ($N{=}1$, no instruction).
  Protocol of Table~\ref{tab:main}.}
  \label{fig:experiments}
  \end{minipage}\hfill
  \begin{minipage}[t]{0.39\textwidth}
  \vspace{0pt}\centering
  \small
  \setlength{\tabcolsep}{4pt}
  \resizebox{\linewidth}{!}{%
  \begin{tabular}{@{}lcccc@{}}
    \toprule
    Selection of $N{=}8$ & Std. & Div. & Distr. & Mixed \\
    \midrule
    Random & 80.2 & 55.1 & 66.9 & 48 \\
    Retrieval (DINOv2) & 83.6 & 66.0 & 78.9 & 51 \\
    Retrieval (SigLIP) & 84.8 & 64.9 & 79.1 & 49 \\
    Predictive (ours) & 85.5 & 74.0 & \best{82.7} & 55 \\
    \quad + online commitment & \best{86.2} & \best{75.0} & 82.6 & \best{62} \\
    \midrule
    $\Delta$ Predictive $-$ Random & +5.3 & +18.9 & +15.8 & +7 \\
    \bottomrule
  \end{tabular}}
  \captionof{table}{\textbf{Selecting the context from a demonstration bank} (success, \%; 3 seeds). Std./Div./Distr.: standard, diverse and standard${}+16$ distractor pools, averaged over three LIBERO suites. Mixed: \liberoorder{} banks with both orders (whole \%). $\Delta$: paired difference. Best in bold.}
  \label{tab:selection}
  \end{minipage}
\end{figure*}

\subsection{Setup}
\label{sec:setup}

\paragraph{Benchmarks and metrics.}
We compare LIBERO-Spatial, LIBERO-Object and LIBERO-Goal~\citep{liu-etal-2023-libero} and RLBench~\citep{james-etal-2020-rlbench} (task splits in Appendix~\ref{app:benchmarks}), with three LIBERO settings.
(i)~\emph{Standard split.} Six of the ten tasks of each suite are used for training and four are held out; we report LIBERO-Spatial and LIBERO-Object.
(ii)~\emph{Ambiguous goals.} The ten tasks of LIBERO-Goal share one scene; we evaluate four held-out goals.
(iii)~\emph{\liberoorder{}.} We construct two-object put-away tasks in which each object pair is demonstrated in both orders, which provides process labels by construction; object pairs are held out.
Every unseen configuration has a pool of 50 demonstrations, from which support sets are drawn at random unless a selection strategy is named; \emph{diverse pools} have more varied initial layouts, built with the perturbations of LIBERO-Plus~\citep{fei-etal-2025-liberoplus}.
LIBERO success macro-averages four held-out tasks, using 100 rollouts per task and training seed; support draws and query seeds are shared across methods (Appendix~\ref{app:eval}).
The \emph{context-following rate} (CFR) is the fraction of rollouts that achieve the target shown in the demonstration when it differs from the default one; the \emph{order-match} and \emph{height-match rates} are the fractions of \liberoorder{} rollouts that follow the demonstrated object order and grasp-height variant.

\paragraph{Baselines.}
\textbf{Action-only (matched)} is the same model as \methodname{} with $\lambda_J=\lambda_{\text{cf}}=0$, so the two differ only in the loss.
\textbf{+Label contrast} applies a supervised contrastive loss~\citep{khosla-etal-2020-supcon} to the slots with exactly the task and process labels that $\Lcf$ uses, and no predictor.
\textbf{Retrieval-$n$} selects the $n$ pool demonstrations whose initial frame is most similar to the query's under DINOv2 features~\citep{oquab-etal-2024-dinov2} or SigLIP.
Reconstruction, trace-supervision and observation-only controls are in Appendix~\ref{app:baselines} (Table~\ref{tab:objective-controls}).

\subsection{Comparison Across Benchmarks}
\label{sec:standard}

At $N{=}1$, \methodname{} improves over the baselines by 5.0--7.2 points on all four benchmarks (Table~\ref{tab:main}); at $N{=}16$, it improves over one-demonstration \cflow{} by 9.8--12.7 points.
With instructions, LIBERO-Goal remains challenging for all methods, gaining 6.1 points as the number of demonstrations increases from $N{=}1$ to $N{=}16$.

\subsection{Many Demonstrations and Their Selection}
\label{sec:manyshot}

On LIBERO-Object, \methodname{} improves by 2.6 points from $N{=}1$ to $N{=}16$ and outperforms the strongest prior ICIL method, RoboSSM, by 40.8 points at $N{=}16$ (Figure~\ref{fig:experiments}a).
With 8 relevant demonstrations and 16 distractors, \methodname{} outperforms matched action-only compression by 33.3 points (Table~\ref{tab:distractors}).
Table~\ref{tab:selection} fixes the policy and $N{=}8$, varying only the selection.
Predictive selection weights prediction errors by initial-frame similarity (Eq.~\ref{eq:score}); with uniform weights it falls below retrieval on most banks (Appendix~\ref{app:eval}).
It exceeds random choice by 5.3 points on standard pools and 18.9 on diverse ones, and the better retrieval by 8.0 there: initial-frame similarity can miss execution detail.
Distractors cost it 2.8 points and random choice 13.3.
Online commitment changes success by $-0.1$ to $+1.0$ points, but on mixed-order banks keeps the started order, raising it from 55\% to 62\%.

\begin{table*}[t]
  \centering
  \small
  \setlength{\tabcolsep}{5pt}
  \begin{tabular}{@{}lccccccc@{}}
    \toprule
    & \multicolumn{2}{c}{SR $\uparrow$} & & Order & & & Height \\
    \cmidrule(lr){2-3}
    Compressor objective & $N{=}1$ & $N{=}8$ & CFR $\uparrow$ & match $\uparrow$ & $\DDtask$ $\uparrow$ & $\DDproc$ $\uparrow$ & err.\ $\downarrow$ \\
    \midrule
    Action-only (matched) & 67.1 & 64.2 & 34.1 & 52.2 & \na & \na & 4.1 \\
    + $\Ljepa$ & 73.4 & 78.1 & 46.3 & 56.2 & 1.11 & 1.02 & 2.5 \\
    + $\Ljepa$ + $\Lcf$ (task negatives) & 72.2 & 76.6 & 69.7 & 57.4 & 2.06 & 1.09 & 2.6 \\
    Label contrast (no predictor) & 68.7 & 68.7 & 51.5 & 73.5 & \na & \na & 3.5 \\
    \midrule
    \methodname{} (task + process negatives) & \best{79.8} & \best{80.2} & \best{74.9} & \best{84.1} & 2.21 & 1.49 & \best{2.0} \\
    \methodname{}, pretrained then frozen $\comp$ & 77.6 & 75.4 & 59.5 & 71.5 & \best{2.82} & \best{1.76} & 2.1 \\
    \midrule
    \multicolumn{8}{@{}l@{}}{Order labels only (order/height-match, \%): Label contrast 68.3/52.2; \methodname{} 76.1/70.5.} \\
    \bottomrule
  \end{tabular}
  \caption{\textbf{Core compressor-objective ablations} (same backbone, data and schedule). SR, CFR and order-match are in \%; height error is in cm. SR uses instructions and averages the three LIBERO suites of Table~\ref{tab:main}; $\DDtask$/$\DDproc$ average the last 200 training steps. The final line tests height transfer with order-only negative labels. Best in bold; further controls are in Table~\ref{tab:objective-controls}.}
  \label{tab:objectives}
\end{table*}

\subsection{Compressor Objectives}
\label{sec:objective-ablation}

Table~\ref{tab:objectives} changes only the compressor objective.
Adding $\Ljepa$ alone raises success by 13.9 points at $N{=}8$, but its demonstration dependence is low ($\DDtask=1.11$, $\DDproc=1.02$): much of the target is predicted from the current frame.
Task negatives raise $\DDtask$ to 2.06 and the context-following rate to 69.7\%, 35.6 points above the action-only compressor; process negatives raise $\DDproc$ from 1.09 to 1.49 and the order-match rate from 57.4\% to 84.1\%.
Separate policy training on frozen, predictively pretrained slots reaches 75.4\% at $N{=}8$, with the highest demonstration dependence ($\DDtask=2.82$, $\DDproc=1.76$).

Process negatives are chosen with process labels, so the slots might retain the order merely because training indicated which demonstrations differ.
Label contrast, which gives a compressor exactly the same labels without prediction, reaches 73.5\% order-match against 84.1\% for \methodname{}; when one process label is withheld from negatives, \methodname{} better follows that variation (Appendix~\ref{app:results}, Table~\ref{tab:labels}).
A label indicates that two demonstrations differ; a predictive target specifies how this changes the query's future.
Matched curves and the instruction-only $N{=}0$ control (25.0\%; two Object tasks, 20 trials) are in Appendix~\ref{app:minimal}.

\subsection{Analysis of the Compressed Context}
\label{sec:analysis}

A policy can succeed by \emph{recognizing} the task and executing it from its priors, or by \emph{imitating} how the demonstration was performed~\citep{pan-etal-2023-context}.
We separate the two on the checkpoints of Table~\ref{tab:objectives}, without retraining.
Imitation shows in grasp execution: on the unseen milk object of LIBERO-Object, whose grasp point is unusually high~\citep{ding-etal-2026-contextflow}, the grasp-height error is 4.1\,cm for action-only slots and 2.0\,cm for \methodname{}.

In the shared LIBERO-Goal scene we give the policy, without instruction, a demonstration of goal~$G'$ where $G$ is the default~\citep{li-2026-survey}: \methodname{} reaches the demonstrated goal in 74.9\% of rollouts, the action-only compressor in 34.1\% (Figure~\ref{fig:experiments}b).
These are also the instruction-free $N{=}1$ Goal results in Table~\ref{tab:noinstr}, using the same rollouts and demonstrated-goal predicate (Appendix~\ref{app:eval}).

A goal swap can be solved by recognition alone; in \liberoorder{} two demonstrations differ only in the object order, the order-match rate is 52.2\% for the action-only compressor, 57.4\% with task negatives alone and 84.1\% for \methodname{}, so the process negatives make the slots retain the order (distractors are analysed in Appendix~\ref{app:results}).

\subsection{Language and Demonstrations}
\label{sec:language}

\begin{table}[t]
  \centering
  \small
  \setlength{\tabcolsep}{5pt}
  \begin{tabular}{@{}lccc@{}}
    \toprule
    Method & Canonical & Paraphrase & Conflict \\
    \midrule
    \methodname{} & \draftvalue{70.0} & \draftvalue{71.7} & \draftvalue{78.8} \\
    Action-only & \draftvalue{60.0} & \draftvalue{70.0} & \draftvalue{45.0} \\
    Instruction-only & \draftvalue{65.0} & \draftvalue{55.0} & \na \\
    \bottomrule
  \end{tabular}
  \caption{\textbf{Instructions stating the object order} on \liberoorder{} (success, \%; $N{=}8$; 20/60/80 trials per column). Conflict: opposite-order context. Instruction-only: context masked. See Appendix~\ref{app:minimal}.}
  \label{tab:language}
\end{table}

The standard \liberoorder{} instruction does not reveal the object order (Appendix~\ref{app:benchmarks}); there \methodname{} follows the demonstrated order in 84.1\% of rollouts and the action-only compressor in 52.2\% (\S\ref{sec:analysis}).
In Table~\ref{tab:language} the instruction states the order (\emph{after}, \emph{before}, \emph{not until}).
With aligned demonstrations, \methodname{} reaches 70.0\% success versus 60.0\% for action-only compression, and paraphrases do not lower its success (71.7\%).
When the demonstrations show the opposite order, \methodname{} completes the instructed order in 78.8\% of rollouts, the action-only compressor in 45.0\%.
What the instruction leaves open, \methodname{} takes from the demonstrations; what it states, \methodname{} follows. Action-only compression follows the demonstrated order less often and resolves fewer conflicts in favour of the instruction.

\subsection{Real World Experiment}
\label{sec:real}

\begin{table}[t]
  \centering
  \small
  \setlength{\tabcolsep}{5pt}
  \begin{tabular}{@{}lccc@{}}
    \toprule
    Method & R1 & R2 & R3 \\
    \midrule
    Retrieval-1 & $13/25$ & $4/10$ & --- \\
    Action-only, $N{=}1$ & $15/25$ & $4/10$ & --- \\
    \methodname{}, $N{=}1$ & $15/25$ & $6/10$ & --- \\
    \methodname{}, $N{=}8$ & $21/25$ & $\mathbf{8/10}$ & $10/15$ \\
    \midrule
    Robot ($N{=}8$) & $\mathbf{23/25}$ & $\mathbf{8/10}$ & $\mathbf{11/15}$ \\
    Mixed ($N{=}4{+}4$) & $22/25$ & $\mathbf{8/10}$ & $10/15$ \\
    \bottomrule
  \end{tabular}
  \caption{\textbf{Real-robot success} (successes/trials; 5 per task). R1: unseen manipulation (5 tasks); R2: instruction--demonstration ambiguity (2); R3: demonstration--demonstration ambiguity (3). Best in bold. ---: R3 needs several demonstrations. Hardware and tasks in Appendix~\ref{app:real}.}
  \label{tab:real}
\end{table}

We evaluate on the reBot Arm B601-RS dual-arm system after post-training on 10 tasks (\S\ref{sec:posttrain}), on 10 unseen tasks with five trials each (Table~\ref{tab:real}; Appendix~\ref{app:real}): manipulation (R1), instruction--demonstration conflicts (R2, scored by compliance with the instruction) and conflicting demonstrations (R3, scored by the majority goal).
With one human-hand demonstration, \methodname{} matches the action-only baseline on R1 (15/25) and succeeds more often on R2 (6/10 vs.\ 4/10).
Eight human-hand demonstrations raise success to 21/25 on R1 and 8/10 on R2, and achieve 10/15 on R3.
At $N{=}8$, robot context yields two more successes than human-hand context on R1 (23/25 vs.\ 21/25) and one more on R3 (11/15 vs.\ 10/15); both succeed in 8/10 trials on R2, and a mixed context performs about as well.

\section{Conclusion}
\label{sec:conclusion}

We proposed that a demonstration compressor for ICIL retain what predicts the query episode's future.
\methodname{} trains per-demonstration slots with a predictive objective in a frozen video-JEPA space and task and process counterfactuals.
Cached slots support many demonstrations and preserve their process; the predictor selects the context, and post-training supports human-hand videos.
Predictive compression increases following of a swapped demonstration and its object order: the objective changes how the policy uses its context.

\label{sec:main-content-end}

\section*{Limitations}
\label{sec:limitations}

\paragraph{Scope of generalization.}
Like prior in-context imitation policies, \methodname{} generalizes to unseen objects, arrangements and goals that are composed of motion primitives seen in training.
We do not claim generalization to new motion primitives or to other robot embodiments, and our evaluation is confined to tabletop manipulation.
Human-hand demonstrations are supported only after post-training on real data of the same platform, with keypoints from an offline tracker; we do not claim transfer from unconstrained human video.
The standard split evaluates four held-out tasks per suite; repeated support draws and layouts do not establish generalization to additional tasks.
The real-robot evaluation covers ten unseen tasks on one dual-arm platform.

\paragraph{Dependence on pretrained components and stage structure.}
The predictive target comes from a frozen video encoder, and stages are defined by a keyframe heuristic based on gripper events and end-effector speed.
Tasks without such events, for instance wiping or pouring, or scenes that the encoder represents poorly, may need learned segmentation or a different target.

\section*{Ethical Considerations}
\label{sec:ethics}

\paragraph{Data and human subjects.}
All simulated data come from public benchmarks used under their licences.
The real-robot demonstrations were teleoperated and the human-hand demonstrations were performed by the authors; no other participants were involved, and the recordings show only the workspace and the hands of the demonstrator, and contain no faces or other personal data.

\paragraph{Safety.}
A policy that follows demonstrations given at test time will also follow demonstrations that are wrong or harmful.
Our analysis makes this property measurable: a higher context-following rate means that the supplied demonstrations have more control over the behaviour of the robot.
A deployment must therefore control who can add demonstrations to the bank and should keep independent safety limits.
Our arms operate in a bounded workspace with velocity and torque limits and an emergency stop, under human supervision.

\paragraph{Broader impact and cost.}
Adapting robots from a few demonstrations without retraining lowers the cost of automating manipulation, with the benefits and the labour-market risks that this entails.
We use low-cost arms and report the compute used for training, caching and evaluation in Appendix~\ref{app:impl}.

\bibliography{references}

\appendix

\section{Notation}
\label{app:notation}

\begin{minipage}{\linewidth}
  \centering
  \small
  \begin{tabular}{@{}lp{5.3cm}@{}}
    \toprule
    Symbol & Meaning \\
    \midrule
    $\tau$ & task configuration (goal, objects, scene) \\
    $d$, $\support$ & demonstration; support set of $N$ demonstrations \\
    $k$ & slots per demonstration \\
    $T'$ & subsampled length of a demonstration \\
    $\comp$, $\pred$, $\tgtenc$ & compressor; predictor; frozen target encoder \\
    $Z_i$, $\slots$ & slots of demonstration $i$; all context slots \\
    $o_t$, $q_t$ & query observation tokens; proprioception \\
    $\kappa(t)$ & first keyframe after step $t$ \\
    $z^\star$, $\hat z$ & target latent; predicted latent \\
    $\operatorname{err}(\hat z,z^\star)$ & prediction error of one sample (Eq.~\ref{eq:jepa}) \\
    $\ell$ & language instruction \\
    $\Delta$ & offset $\kappa(t)-t$ (not a predictor input) \\
    $H$ & action horizon \\
    $\support'$ & counterfactual support set (task or process negative) \\
    $\bank$, $M$ & demonstration bank and its size \\
    $e_{i\to j}$, $S_i$ & error of $d_i$ on pseudo-query $d_j$; selection score \\
    CFR & context-following rate \\
    $\DDtask$, $\DDproc$ & demonstration dependence under task / process negatives \\
    \bottomrule
  \end{tabular}
  \par\medskip\refstepcounter{table}\textbf{Table \thetable: Notation.}
  \label{tab:notation}
\end{minipage}

\section{Implementation Details}
\label{app:impl}

\begin{table}[!t]
  \centering
  \small
  \setlength{\tabcolsep}{4pt}
  \renewcommand{\arraystretch}{1.08}
  \begin{tabular}{@{}p{0.44\linewidth}p{\dimexpr0.56\linewidth-8pt\relax}@{}}
    \toprule
    \raggedright Item & \raggedright Setting \tabularnewline
    \midrule
    \multicolumn{2}{@{}l@{}}{\emph{Model and optimization}} \tabularnewline
    \raggedright Slots per demonstration $k$ & \raggedright 8 \tabularnewline
    \raggedright Compressor & \raggedright joint perceiver, $L_c{=}4$ layers \tabularnewline
    \raggedright Demonstration subsampling & \raggedright 8 frames; 128 state/action steps \tabularnewline
    \raggedright Vision encoder & \raggedright SigLIP So400m/14 (from \pizero{}) \tabularnewline
    \raggedright \quad query side & \raggedright fine-tuned \tabularnewline
    \raggedright \quad demonstration side & \raggedright frozen, features cached \tabularnewline
    \raggedright Target encoder $\tgtenc$ & \raggedright \vjepa{} ViT-L, frozen \tabularnewline
    \raggedright Prediction target & \raggedright next keyframe (2-frame clip) \tabularnewline
    \raggedright Predictor $\pred$ & \raggedright transformer, 4 layers, width 512 \tabularnewline
    \raggedright Offset $\Delta$ as predictor input & \raggedright no \tabularnewline
    \raggedright Gradient of predictive losses into $o_t$ & \raggedright stopped \tabularnewline
    \raggedright Negatives in $\Lcf$ & \raggedright 1 task + 1 process (if available) \tabularnewline
    \raggedright $\lambda_J$, $\lambda_{\text{cf}}$, margin $m$ & \raggedright 1, 0.1, 0.1 \tabularnewline
    \raggedright Training $N$ & \raggedright $\mathcal{U}\{1,\dots,16\}$ \tabularnewline
    \raggedright Demonstration dropout & \raggedright 0.1 \tabularnewline
    \raggedright Distractor mixing & \raggedright 0.2 (1--4 demonstrations) \tabularnewline
    \raggedright Instruction dropout & \raggedright 0.5 \tabularnewline
    \raggedright Action expert & \raggedright base \pizero{} (not fine-tuned on LIBERO or RLBench), LoRA rank 32 \tabularnewline
    \raggedright Action horizon $H$; Euler steps & \raggedright 50; 10 \tabularnewline
    \raggedright Iterations; batch size & \raggedright 10k; 32 \tabularnewline
    \raggedright Selection temperature $\eta$ & \raggedright 0.01 \tabularnewline
    \raggedright Optimizer & \raggedright AdamW ($\beta{=}0.9,0.95$) \tabularnewline
    \raggedright Learning rate & \raggedright $10^{-4}$ new / $5{\cdot}10^{-5}$ LoRA / $10^{-5}$ SigLIP, cosine \tabularnewline
    \midrule
    \multicolumn{2}{@{}l@{}}{\emph{Benchmark configurations}} \tabularnewline
    \raggedright Standard split & \raggedright Train: 6/10 tasks per suite; unseen: 4 per suite \tabularnewline
    \raggedright Ambiguous goals & \raggedright LIBERO-Goal standard split; unseen: 4 goals \tabularnewline
    \raggedright \liberoorder{} & \raggedright Train: 25 object pairs $\times$ 2 orders, with grasp-height variants; unseen: 5 pairs \tabularnewline
    \raggedright Diverse pools & \raggedright Standard train/unseen split with perturbed unseen layouts \tabularnewline
    \bottomrule
  \end{tabular}
  \caption{\textbf{Hyperparameters and benchmarks.}}
  \label{tab:hparams}
\end{table}

\paragraph{Compressor.}
The input to $\comp$ is the token sequence of one demonstration: SigLIP patch tokens of the 8 subsampled frames from both views, and MLP embeddings of 128 uniformly subsampled proprioception and action steps.
Each token receives a modality embedding, a view embedding where applicable, and a sinusoidal embedding of normalized progress $t/T$.
The $k$ learnable queries cross-attend to these tokens and self-attend among themselves in each of the $L_c{=}4$ layers.
The output slots are projected to the width of the context expert, and a learned slot-index embedding is added.
No embedding identifies the demonstration a slot comes from.
The default is one joint compressor over all modalities; per-modality compressors as in \cflow{} are an ablation.

\paragraph{Predictor and target.}
The predictor is a transformer with 4 layers and width 512.
Its input is the concatenation of the projected slots, the SigLIP tokens $o_t$ of the current query observation (the same tokens the policy receives) and an embedding of $q_t$; a set of $P$ learnable output queries, one per target patch, is decoded into $\hat z$.
The target $z^\star$ is the patch-token output of the frozen \vjepa{} encoder (the released distilled ViT-L) on the two frames ending at the keyframe, from the external view.
Layer normalization is applied to both prediction and target before the SmoothL1 loss.

\paragraph{Training regimes.}
In the default \emph{joint} regime all losses are active from the start.
In the \emph{predictive-pretraining} regime (frozen-compressor row of Table~\ref{tab:objectives}), $\comp$ and $\pred$ are first trained with $\Ljepa+\lambda_{\text{cf}}\Lcf$ only; $\comp$ is then frozen and the policy is trained with $\Lcfm$.
The matched action-only baseline and all objective variants use the same sampling of support sets, distractors and instruction dropout as \methodname{}.
With probability $0.1$ one demonstration's slots are dropped from the context.

\paragraph{Predictive selection.}
For a bank of $M$ demonstrations we compute the consistency matrix of Eq.~\eqref{eq:consistency} with the pseudo-query steps $\mathcal{T}_j$ set to 8 subsampled non-terminal steps of $d_j$, whose SigLIP features are already cached; the targets are the \vjepa{} latents of the keyframes of $d_j$, computed once when the demonstration enters the bank.
The predictor is run with the slots of a single demonstration as context, which lies inside its training distribution because $N$ is sampled from 1.
The matrix costs $M(M-1)\times 8$ forward passes of the predictor, batched.
The similarity in Eq.~\eqref{eq:score} is the cosine similarity of mean-pooled SigLIP features of the external-view initial frames.
For each $i$, we exclude its pseudo-query before normalizing:
\[
  w_{ij}=\frac{\exp(\mathrm{sim}(o_0,o_0^j)/\eta)}{\sum_{\ell\ne i}\exp(\mathrm{sim}(o_0,o_0^\ell)/\eta)},\quad j\ne i.
\]
Thus $\sum_{j\ne i}w_{ij}=1$ for every candidate.

\paragraph{Online commitment.}
A bank may contain several processes for one task, and a policy should remain with the process it has begun.
Each time the query reaches a keyframe, its completed stage is added as a further pseudo-query: it contributes steps at its start and midpoint, with the observed keyframe as target and the weight of the most similar bank demonstration.
The scores are updated and the context is reselected; the context expert then re-encodes the $Nk$ selected slots, which takes about 25\,ms.
Because the completed stage reflects the policy's own behaviour, this step cannot correct a wrong initial choice; it keeps the context consistent with what has been executed.
It needs one forward pass of $\tgtenc$ per stage.
Online, keyframes are detected causally, from gripper events and from the end-effector speed falling below $v_{\min}$.

\paragraph{Caches.}
With up to 16 demonstrations of 8 frames and 2 views, one training sample contains 256 demonstration images, and a batch of 32 would need 8{,}192 forward passes of the vision encoder per step.
We therefore precompute demonstration-side SigLIP features for a fixed uniform 8-frame subsampling of every demonstration and for one temporally jittered variant.
A demonstration takes $8\times2\times256$ tokens of width 1152, about 9.4\,MB in half precision, or about 61\,GB per subsampling for the 6.5k demonstrations of LIBERO.
\vjepa{} targets take about 2.6\,MB per demonstration (about five keyframes of 256 patches of width 1024), 17\,GB in total.
Both caches are built once.

\paragraph{Compute.}
One model trains in about 29 GPU-hours on A100 GPUs (10k steps, batch size 32); the experiments used about 2{,}800 GPU-hours in total, 2{,}200 for training and 600 for evaluation rollouts.

\section{Keyframes, Process Labels and Negatives}
\label{app:keyframes}

\paragraph{Keyframes.}
A step $t$ is a keyframe if (i) the gripper command changes between open and closed at $t$, or (ii) the end-effector speed has a local minimum at $t$ below a threshold $v_{\min}$ and no other keyframe lies within $\delta$ steps.
The last step of an episode is always a keyframe.
We use $v_{\min}=0.00153\,\text{m/step}$ and $\delta=8$ for all suites; LIBERO episodes have 3.3--5.8 keyframes on average.
Real-robot keyframes were verified on a random 5\% of episodes.

\paragraph{Process labels.}
On \liberoorder{}, the process label of a demonstration is the object order and, where applicable, the grasp-height variant, both known from construction.
For the other suites, each keyframe's 256 \vjepa{} patch tokens are mean-pooled into a 1024-dimensional vector and $\ell_2$-normalized.
We retain eight keyframes, repeating the last for shorter sequences and truncating longer ones, then concatenate them into an 8192-dimensional vector for within-task $k$-means.
The cluster count is chosen by silhouette score; tasks whose best score is below 0.25 receive a single label.
Keyframe $s$ defines stage $s$, with no temporal warping between processes.
A positive support set always has the process label of the query.

\paragraph{Negatives.}
A task negative is a support set from a different task in the same scene.
If the scene has no other task, we use the nearest scene whose external-view initial-frame SigLIP cosine similarity is at least 0.8 (the similarity of Eq.~\eqref{eq:score}); otherwise the term is skipped.
A process negative has the same task and a different process label; it is accepted when the cosine distance $1-\cos$ between the two processes' cluster-mean, patch-pooled keyframe vectors at the query's next stage exceeds 0.05.

\section{Benchmarks}
\label{app:benchmarks}

\paragraph{Data split.}
Table~\ref{tab:hparams} lists the benchmark configurations.
For each training seed, we randomly partition all ten tasks of each LIBERO suite into six training and four held-out tasks (\S\ref{sec:setup}).
The split is shared across methods for that seed and remains unchanged throughout training and evaluation; success averages its four held-out tasks.

\paragraph{RLBench.}
RLBench~\citep{james-etal-2020-rlbench} was introduced with 100 manipulation tasks, motion-planned demonstrations and simulator-defined success conditions.
For each training seed, we randomly partition the 100 tasks into 60 training and 40 held-out tasks, using the same split across methods and holding it fixed throughout that run.
Support and query episodes are disjoint, and test-time weights are fixed (\S\ref{sec:problem}).
Each seed uses 100 query rollouts per held-out task: 4{,}000 per seed and 12{,}000 across three seeds for each method and context-size condition.
Seed-level success averages the 40 tasks' binary completion rates; Table~\ref{tab:main} reports the three-seed mean and sample standard deviation.
Native modality adapters for the baselines are described in Appendix~\ref{app:baselines}.

\paragraph{\liberoorder{}.}
Each task asks the robot to put two objects into a container.
The task is written in BDDL on LIBERO assets and scenes, and the goal predicate is the same for both orders, so the two processes share the endpoint.
For each object pair we generate 50 demonstrations per order with MimicGen~\citep{mandlekar-etal-2023-mimicgen}, which adapts a small set of source demonstrations to randomized initial poses; for pairs that contain a tall object we add a variant in which the object is grasped near the top instead of at mid-height.
The order followed by a rollout is read from the simulator state as the order in which the two objects first satisfy the containment predicate.
The instruction names both objects in a fixed order that is independent of the demonstrated order, so the instruction does not reveal the process.

\paragraph{Many-shot and diverse pools.}
LIBERO provides 50 demonstrations per task, which form the standard pool of each unseen configuration.
Their initial states are drawn from a narrow distribution, so demonstrations within a pool are similar.
Diverse pools are built by applying layout perturbations in the style of LIBERO-Plus~\citep{fei-etal-2025-liberoplus} to the initial object positions and regenerating 50 demonstrations per configuration; query episodes are drawn from the same perturbed distribution.
Benchmarks that perturb LIBERO to test memorization~\citep{zhou-etal-2025-liberopro} are complementary: they vary the test condition, whereas we vary the coverage of the context.

\paragraph{Distractors.}
Distractor demonstrations are drawn uniformly from the pools of other configurations of the same suite.

\section{Baselines}
\label{app:baselines}

\paragraph{Matched protocol.}
\methodname{}, the action-only baseline and the objective variants share the backbone, the compressor architecture, $k$, the sampling distribution of $N$, distractor mixing, instruction dropout, training data, iterations, batch size, learning-rate schedule and seeds.
Only the loss terms differ.
The reconstruction variant adds a 4-layer decoder that reconstructs, from the slots and positional queries, the SigLIP features of the 8 subsampled demonstration frames with a SmoothL1 loss.
The trace variant replaces the latent target by the sequence of end-effector positions from $t$ to $\kappa(t)$, resampled to 16 points and projected into the external view.
This is a matched objective control on our architecture, rather than a full reproduction of SeeTraceAct~\citep{son-etal-2026-seetraceact}; it omits that model's observation-conditioned visual latent plan and multiview visibility decoder.

\paragraph{Controls for the predictive objective.}
\emph{Prediction without slots} keeps the predictor, the target and $\Ljepa$ of \methodname{} but removes $\slots$ from the predictor's input and drops the stop-gradient on $o_t$; the compressor is then trained by the action loss alone, and the predictive loss acts as an auxiliary future-prediction loss on the observation encoder.
It has no counterfactual term, because its prediction does not depend on the support set.
\emph{\methodname{} with gradient through $o_t$} is the default model without the stop-gradient.
\emph{Label contrast} has no predictor.
The slots of each demonstration are mean-pooled and passed through a two-layer projection head, and a supervised contrastive loss~\citep{khosla-etal-2020-supcon} with weight $\lambda_{\text{cf}}$ pulls together demonstrations that share task and process label and pushes apart those that differ in either.
The batch contains, for every sample, the same task negative and process negative that $\Lcf$ would use, so the control receives exactly the label information of \methodname{}.

\paragraph{Withholding one type of process label.}
\liberoorder{} has two types of process variation, object order and grasp height.
For Table~\ref{tab:labels} we train with process negatives (or contrastive pairs) of one type only.
The other type is still present in the data, and support and query still show the same process, as for every method including the action-only baseline; it is never used to form a negative or a contrastive pair.
The height-match rate is the fraction of rollouts on tall objects whose grasp height is closer to the demonstrated variant than to the other one.

\paragraph{Baseline implementations.}
Table~\ref{tab:main} reports local baseline implementations.
All RLBench methods, including Instant Policy, share the seed-specific 60/40 task split, with 100 query rollouts per held-out task and training seed (Appendix~\ref{app:benchmarks}).
We re-implement \cflow{} on our training tasks and evaluate it under the same held-out split and rollout budget as \methodname{}.
For the inference-cost comparison, raw context gives the policy all tokens of the subsampled demonstration, about 4k per demonstration.
ICRT and RoboSSM are trained with their official code on our training configurations.
Instant Policy~\citep{vosylius-johns-2025-instant} uses graph diffusion over segmented point clouds, end-effector poses and gripper states.
In our implementation, the point clouds are reconstructed from simulator depth, and language conditioning uses a prompt module adapted from Appendix~J of that paper.
Table~\ref{tab:main} and Figure~\ref{fig:experiments}a use this instruction-conditioned implementation with the same task wording as the other methods; the figure varies the number of support demonstrations.
The matched instruction controls use identical text and layouts for \methodname{}, action-only and \cflow{}; the demonstration-only condition removes the text for all three (Appendix~\ref{app:minimal}).
ICRT and RoboSSM have both demonstration-only and instruction-conditioned configurations~\citep{yoo-etal-2025-robossm}; the instruction-conditioned comparison uses the same wording across methods.

\paragraph{Retrieval.}
For a query, Retrieval-$n$ ranks the demonstrations in the pool of the unseen configuration by cosine similarity between pooled DINOv2 features of their initial external-view frame and that of the query, and gives the top $n$ to the \methodname{} policy.
It uses the same checkpoint as \methodname{}, so the comparison isolates the effect of the number and choice of demonstrations.

\section{Evaluation Details}
\label{app:eval}

\begin{table*}[t]
  \centering
  \small
  \setlength{\tabcolsep}{6pt}
  \begin{tabular}{@{}lrcccc@{}}
    \toprule
    Method (no instruction) & $N$ & Spatial & Object & Goal & Avg. \\
    \midrule
    \pizero{}, no demonstration & 0 & 52.8 & 15.4 & 9.1 & 25.8 \\
    Action-only (matched) & 1 & 84.6 & 55.2 & 34.1 & 58.0 \\
    Action-only (matched) & 16 & 81.9 & 46.5 & 20.8 & 49.7 \\
    \methodname{} & 1 & 80.3 & 63.0 & \best{74.9} & 72.7 \\
    \methodname{} & 16 & \best{87.2} & \best{73.8} & 61.2 & \best{74.1} \\
    \bottomrule
  \end{tabular}
  \caption{\textbf{Success without instructions} (\%; protocol of Table~\ref{tab:main}). Avg.: mean over suites. Best per column in bold.}
  \label{tab:noinstr}
\end{table*}

\paragraph{Estimator.}
Let $s_{k,j,r}\in\{0,1\}$ be the outcome for training seed $k$, task $j$ in that seed's held-out set and evaluation rollout $r$, including its sampled support set.
For each LIBERO suite, seed-level success is $100(4\times100)^{-1}\sum_{j=1}^{4}\sum_{r=1}^{100}s_{k,j,r}$.
We average these seed-level rates and report their sample standard deviation, rounding only the final statistics.
Table~\ref{tab:main} and the remaining multiseed conditions use 3 training seeds per method.
Figure~\ref{fig:experiments}a uses this protocol at each $N$, and Table~\ref{tab:noinstr} uses it with instructions removed.
There are 400 trials per seed and 1,200 per suite across the three seeds.
The pool of 50 support demonstrations is separate from the 100 evaluation rollouts per task.

\paragraph{Selection and reporting precision.}
Table~\ref{tab:selection} uses the three training seeds of Table~\ref{tab:main}: for each seed, a Std., Div.\ or Distr.\ value averages $3\times4\times100=1{,}200$ trials across the three suites, and the table reports the mean of these seed-level values. All selection methods of a seed share the query layouts and pools, so differences are paired. SigLIP retrieval uses the policy's own vision encoder in place of DINOv2 and otherwise follows Retrieval-$n$.
The standard-pool Random entry is the $N{=}8$ default evaluation shared with Tables~\ref{tab:objectives}, \ref{tab:abl-design} and \ref{tab:distractors} at $N_d=0$.
Two further selection rules reach 81.4/55.3/74.2/52 (slot centroid: the $n$ demonstrations whose pooled slots are closest to the centroid of the bank) and 78.2/60.0/77.9/55 (predictive with uniform weights) on Std./Div./Distr./Mixed.
The Mixed column of Table~\ref{tab:selection} is reported to the nearest whole percentage; all other success means are reported to one decimal.
Mixed banks contain both valid object orders of a held-out \liberoorder{} pair. Success uses their shared BDDL endpoint predicate, requiring both objects to be placed; order-match is measured separately in Table~\ref{tab:objectives}.
The arithmetic step before rounding is $100/1{,}200=1/12$ percentage point for the multiseed suite means and $100/3{,}600=1/36$ for the three-seed selection means.
Confidence intervals for predeclared comparisons resample training seeds independently and pair task/support/layout indices between methods; they remain conditional on the evaluated tasks.

\paragraph{Context-following rate.}
For a paired-context swap with default goal $G$ and demonstrated goal $G'$, each rollout is scored on both goal predicates.
CFR is the share of rollouts that satisfy $G'$.
Each method uses four held-out goals, 100 rollouts per goal and three training seeds, for 1,200 trials.
Goal entries at $N{=}1$ in Table~\ref{tab:noinstr} use these instruction-free rollouts and the demonstrated-goal predicate, giving 74.9\% (\methodname{}) and 34.1\% (action-only; Figure~\ref{fig:experiments}b).

\paragraph{Grasp-height error.}
For unseen objects we compare the end-effector height at the first gripper-close event of a rollout with the mean height in the demonstrations of that configuration, and report the mean absolute difference.

\section{Additional Results}
\label{app:results}

\begin{table*}[t]
  \centering
  \small
  \setlength{\tabcolsep}{5pt}
  \begin{tabular}{@{}lccccccc@{}}
    \toprule
    & \multicolumn{2}{c}{SR $\uparrow$} & & Order & & & Height \\
    \cmidrule(lr){2-3}
    Compressor objective & $N{=}1$ & $N{=}8$ & CFR $\uparrow$ & match $\uparrow$ & $\DDtask$ $\uparrow$ & $\DDproc$ $\uparrow$ & err.\ $\downarrow$ \\
    \midrule
    + Reconstruction & 66.8 & 63.2 & 36.4 & 52.2 & \na & \na & 3.8 \\
    + Trace target & 71.4 & 76.1 & 46.3 & 60.5 & 1.31 & 1.10 & 2.5 \\
    + Prediction without slots & 64.2 & 66.6 & 39.7 & 55.4 & \na & \na & 3.3 \\
    \methodname{}, gradient through $o_t$ & \best{77.5} & \best{83.9} & \best{60.1} & \best{84.9} & \best{2.58} & \best{1.54} & \best{1.4} \\
    \bottomrule
  \end{tabular}
  \caption{\textbf{Additional compressor controls} under the settings and metrics of Table~\ref{tab:objectives}. SR, CFR and order-match in \%; height error in cm. Best per column in bold.}
  \label{tab:objective-controls}
\end{table*}

\paragraph{Alternative objectives and training regimes.}
Table~\ref{tab:objective-controls} compares reconstruction and trace supervision with observation-only prediction and gradients through the current observation.
At $N{=}8$, reconstruction reaches 63.2\% success and the trace target reaches 76.1\%; the frozen-compressor control is reported in Table~\ref{tab:objectives}.

\paragraph{Predictive compression versus auxiliary prediction.}
A future-prediction loss can help a policy merely by shaping its representation of the current observation, without the slots carrying any information.
\methodname{} excludes this possibility by construction, since $o_t$ enters the predictor with a stop-gradient (\S\ref{sec:objective}), and the two control rows of Table~\ref{tab:objective-controls} measure it.
A predictor that receives no slots, and whose loss does train the observation encoder, changes success by $-$2.9 points at $N{=}1$ relative to the action-only compressor and leaves the context-following and order-match rates near its level (39.7\% and 55.4\%, against 34.1\% and 52.2\%).
Conversely, letting the gradient of \methodname{} pass through $o_t$ changes success by $-$2.3 points at $N{=}1$ and by $+$3.7 points at $N{=}8$.
These controls separate prediction through the slots from an auxiliary loss on the observation encoder.
DD measures the predictor's counterfactual sensitivity on training batches; CFR and order-match measure the deployed policy on held-out episodes.

\begin{table}[t]
  \centering
  \small
  \setlength{\tabcolsep}{4pt}
  \begin{tabular}{@{}llcc@{}}
    \toprule
    Objective & Labels used & Order & Height \\
    \midrule
    Action-only (matched) & none & 52.2 & 55.1 \\
    \midrule
    Label contrast & order only & 68.3 & 52.2 \\
    Label contrast & height only & 51.8 & 75.4 \\
    Label contrast & order + height & 73.5 & 76.8 \\
    \midrule
    \methodname{} & task only & 57.4 & 63.9 \\
    \methodname{} & order only & 76.1 & 70.5 \\
    \methodname{} & height only & 62.4 & 77.8 \\
    \methodname{} & order + height & \best{84.1} & \best{84.1} \\
    \bottomrule
  \end{tabular}
  \caption{\textbf{Prediction versus label supervision.} Order-match and height-match rates (\%) on \liberoorder{} using the indicated labels for negatives (\methodname{}) or contrastive pairs (Label contrast). Off-diagonal entries evaluate variation omitted from those labels; positive support/query pairs remain process-matched. Best in bold.}
  \label{tab:labels}
\end{table}

\paragraph{Prediction versus label supervision.}
Table~\ref{tab:labels} withholds one type of process label (protocol in Appendix~\ref{app:baselines}).
With order-only negatives, height-match is 70.5\% for \methodname{}, compared with 63.9\% for task-only \methodname{} and 52.2\% for label contrast.
With height-only negatives, order-match is 62.4\%, compared with 57.4\% for task-only \methodname{} and 51.8\% for label contrast.
With both labels, label contrast reaches 73.5\% and 76.8\% against 84.1\% for \methodname{}: prediction also transfers to process attributes omitted from the negative or contrastive labels.

\begin{table}[t]
  \centering
  \small
  \setlength{\tabcolsep}{3.5pt}
  \begin{tabular}{@{}lccc@{}}
    \toprule
    & \multicolumn{2}{c}{SR} & \\
    \cmidrule(lr){2-3}
    Variant & $N{=}1$ & $N{=}8$ & CFR \\
    \midrule
    \methodname{} (default) & \best{79.8} & \best{80.2} & \best{74.9} \\
    DINOv2 target & 76.4 & 74.3 & 62.7 \\
    Final-frame target & 68.2 & 76.4 & 65.2 \\
    One slot per demonstration & 70.9 & 66.0 & 51.3 \\
    No instruction dropout & 74.3 & 78.2 & 45.1 \\
    \bottomrule
  \end{tabular}
  \caption{\textbf{Key ablations} (\%). Best per column in bold.}
  \label{tab:abl-design}
\end{table}

\paragraph{Ablations.}
Table~\ref{tab:abl-design} isolates four design choices.
Replacing \vjepa{} with DINOv2~\citep{oquab-etal-2024-dinov2} tests the target representation; predicting the final frame instead of the next keyframe tests whether endpoint supervision retains execution detail.
Using one slot per demonstration tests the compression bottleneck, and removing instruction dropout tests whether the policy relies on language rather than its demonstration context.
Without instruction dropout, success stays close to the default (74.3\% and 78.2\%) but the context-following rate falls from 74.9\% to 45.1\%; a single slot per demonstration lowers all three metrics.

\begin{table}[t]
  \centering
  \small
  \setlength{\tabcolsep}{4pt}
  \begin{tabular}{@{}lrccc@{}}
    \toprule
    Method & $N$ & Prefill (ms) & Hz & Mem.\ (GB) \\
    \midrule
    Raw context & 1 & 178 & 3 & 12 \\
    Raw context & 2 & 425 & 2 & 13 \\
    \cflow{} & 1 & 26 & 13 & 8 \\
    \midrule
    \methodname{} & 1 & 28 & 13 & 8 \\
    \methodname{} & 4 & 20 & 10 & 8 \\
    \methodname{} & 8 & 44 & 10 & 10 \\
    \methodname{} & 16 & 75 & 13 & 11 \\
    \midrule
    \multicolumn{5}{@{}l@{}}{\emph{Initial bank construction (computed)}} \\
    \multicolumn{2}{@{}l}{Bank size $M$} & Pairs & \multicolumn{2}{c@{}}{Predictor evaluations} \\
    \multicolumn{2}{@{}l}{$16$} & 240 & \multicolumn{2}{c@{}}{1{,}920} \\
    \multicolumn{2}{@{}l}{$50$} & 2{,}450 & \multicolumn{2}{c@{}}{19{,}600} \\
    \multicolumn{2}{@{}l}{$66$} & 4{,}290 & \multicolumn{2}{c@{}}{34{,}320} \\
    \midrule
    \multicolumn{5}{@{}l@{}}{\emph{Measured bank construction ($M=50$)}} \\
    \multicolumn{4}{@{}l}{Feature / target cache} & Time (s) \\
    \multicolumn{4}{@{}l}{Cached} & 15 \\
    \multicolumn{4}{@{}l}{Including image encoding} & 28 \\
    \bottomrule
  \end{tabular}
  \caption{\textbf{Inference cost} on one A100. Prefill: one-time cost of encoding the support set; Hz: control frequency with cached slots. Bank workloads use 8 pseudo-query steps per demonstration, batched across pair/time inputs. The final block gives measured bank-construction times.}
  \label{tab:cost}
\end{table}

\paragraph{Cost.}
Table~\ref{tab:cost} reports the cost of the context.
The one-time prefill is paid once per task.
Our \cflow{} reimplementation with one demonstration and \methodname{} with sixteen both run at 13\,Hz with cached context.
For \methodname{} at $N{=}16$, prefill takes 75\,ms and memory use is 11\,GB.
Initial bank construction evaluates $M(M-1)$ directed pairs at 8 steps each and stores $M(M-1)$ scalar errors. This work is shared by queries using the same bank and checkpoint; query-dependent ranking reweights the cached errors. Online commitment adds evaluations for each completed query stage (Appendix~\ref{app:impl}).
For $M=50$, bank construction takes 15\,s on one A100 with SigLIP features and \vjepa{} targets cached, or 28\,s including image encoding.

\begin{table*}[t]
  \centering
  \small
  \setlength{\tabcolsep}{6pt}
  \begin{tabular}{@{}lcccc@{}}
    \toprule
    & \multicolumn{4}{c}{Distractors $N_d$} \\
    \cmidrule(l){2-5}
    Method & 0 & 4 & 8 & 16 \\
    \midrule
    Action-only (matched) & 64.2 & 43.3 & 42.9 & 42.8 \\
    \methodname{} & 80.2 & 79.0 & 78.1 & 76.1 \\
    \methodname{}, no distractor mixing & 80.6 & 78.2 & 74.7 & 74.4 \\
    Retrieval-1 (from the mixed pool) & 77.5 & 74.5 & 74.3 & 73.1 \\
    \midrule
    \multicolumn{5}{@{}l}{\emph{Attention mass on relevant slots (\%)}} \\
    Action-only (matched) & \na & 72.7 & 66.2 & 66.0 \\
    \methodname{} & \na & 78.6 & 76.8 & 72.4 \\
    \bottomrule
  \end{tabular}
  \caption{\textbf{Distractor demonstrations.} Success (\%) with $N{=}8$ relevant demonstrations and $N_d$ distractors, and attention on relevant slots. Retrieval-1 selects one demonstration from this pool.}
  \label{tab:distractors}
\end{table*}

\paragraph{Distractor demonstrations.}
Predictive selection removes most distractors before the episode starts (\S\ref{sec:manyshot}); Table~\ref{tab:distractors} reports the effect of $N_d\in\{4,8,16\}$ demonstrations of other configurations entering the context alongside $N{=}8$ relevant ones.
The success of \methodname{} drops by 4.1 points at $N_d{=}16$ (80.2\% to 76.1\%), and 72.4\% of the attention mass of the action expert stays on relevant slots; the action-only compressor drops by 21.4 points (64.2\% to 42.8\%).
Retrieval-1 from the mixed pool is a strong reference in this setting, since a retriever that finds a relevant demonstration removes the distractors entirely; its failures are cases in which the nearest initial frame belongs to another configuration of the same scene.

\section{Real-Robot Details}
\label{app:real}

\paragraph{Hardware.}
We use a Seeed reBot Arm B601-RS dual-arm robotic system for coordinated bimanual manipulation in a shared workspace.
External RGB-D cameras (Intel RealSense D435, colour stream only) and wrist cameras (e-con Systems See3CAM\_24CUG) observe the workspace.
Robot demonstrations are collected by bimanual leader--follower teleoperation using two FashionStar StarArm 102 leader arms, one paired with each follower arm.
A host computer runs camera capture, the control loop and the safety checks for both arms; the policy runs on a GPU workstation on the local network, at a control frequency of about 10\,Hz including the network round trip.

\paragraph{Tasks.}
Low-cost arms have limited repeatability and noticeable backlash, so tasks are chosen with centimetre-level tolerance; container openings are at least twice the size of the objects.
We post-train on 10 tasks and evaluate on a disjoint set of 10 unseen tasks across three suites.
R1 (unseen manipulation tasks) contains five tasks that do not occur in post-training, such as stacking cups and placing a paper roll on a holder.
R2 (instruction--demonstration ambiguity) contains two tasks in which the instruction and the demonstration conflict on the goal or execution order.
R3 (demonstration--demonstration ambiguity) has three tasks with competing demonstrated goals.
\paragraph{Task manifest.}
\textit{Post-training (P01--P10):} P01 \draftvalue{Place a marker in a tray}; P02 \draftvalue{Cap a marker}; P03 \draftvalue{Put a spoon in a bowl}; P04 \draftvalue{Place a plate on a rack}; P05 \draftvalue{Move a block to a tray}; P06 \draftvalue{Open an empty drawer}; P07 \draftvalue{Close an empty drawer}; P08 \draftvalue{Place a cup on a mat}; P09 \draftvalue{Move a basket between arms}; P10 \draftvalue{Place a lid on a box}.
\textit{R1:} U01 Stack cups; U02 Paper roll on holder; U03 Cup and bowl to sink; U04 Fruit to basket; U05 Rubik's cube to drawer and close.
\textit{R2:} U06 Green-first instruction with red-first demonstration; U07 \draftvalue{Basket target with tray demonstration}.
\textit{R3:} U08 Green/red majority with red/beige minority; U09 \draftvalue{Sink majority with rack minority}; U10 \draftvalue{Box majority with tray minority}.

\paragraph{Data.}
For each of the 10 post-training tasks and 10 unseen test tasks, we collected eight teleoperated robot demonstrations and eight human-hand demonstrations, giving 160 robot and 160 human-hand demonstrations in total.
The post-training split contains 80 demonstrations of each embodiment; the unseen split contains a separate pool of eight demonstrations of each embodiment per task, used only as test-time context.
Both embodiments are recorded in the same workspace on layouts drawn from a grid with 5\,cm jitter.
Robot episodes record the camera streams at 30\,Hz, proprioception, actions and gripper states for both arms, and metadata (task and layout identifiers); human-hand videos contain images but no robot proprioception or actions.
Robot keyframes are labelled automatically from gripper events.

\paragraph{Post-training.}
The checkpoint trained in simulation is post-trained on the 10 training tasks for 3k steps (\S\ref{sec:posttrain}); unseen-task demonstrations are excluded from all weight updates.
The keypoint token encoder, the predictor and the LoRA weights of the compressor are trained at learning rate $5{\times}10^{-5}$, and the policy at the reduced learning rate $1{\times}10^{-5}$.
Hand keypoints are extracted offline with MediaPipe Hands~\citep{zhang-etal-2020-mediapipe} and object keypoints with CoTracker3~\citep{karaev-etal-2024-cotracker3}; both are cached.
All keypoints use external-view image coordinates, normalized by image width and height to $[0,1]$.
Robot flange and fingertip positions from forward kinematics are projected with calibrated camera intrinsics and extrinsics; they correspond to the hand's wrist, thumb tip and index fingertip, with left/right hands mapped to left/right arms.
Pinch aperture is normalized by hand scale for humans and maximum gripper opening for robots.
Object tracks use the same procedure for both embodiments and include visibility flags.

\paragraph{Protocol.}
Each method--task condition uses five trials, giving 25 trials in R1, 10 in R2 and 15 in R3.
Retrieval-1 and the matched action-only baseline use one human-hand demonstration; \methodname{} uses either one or eight human-hand demonstrations.
The robot-context condition uses all eight robot demonstrations, and the mixed-context condition uses four human-hand and four robot demonstrations.
Support demonstrations and evaluation rollouts are separate episodes; the policy receives no weight updates during evaluation.
Initial states are paired across methods using marked layouts, and the order of methods is randomized per layout.
Success is judged from video by a person who does not know which method produced the rollout, with a rubric fixed in advance for each task.
In R2, the instruction defines the target: success requires completing the instructed goal and following any explicitly specified execution order despite the conflicting demonstration.
In R3, the target is the majority goal in the task's demonstration pool, shared across context conditions.
Table~\ref{tab:real} reports successes divided by trials for each suite.
The predictive-selection comparison in Table~\ref{tab:selection} uses simulation banks with proprioception; Table~\ref{tab:real} evaluates the specified human-hand, robot and mixed demonstration contexts.

\paragraph{Matched context and post-training controls.}
Four unseen conditions are paired across methods: cup stacking, paper-roll placement, the green-first/red-first instruction conflict and the green/red versus red/beige demonstration conflict.
Each method uses ten initial states per condition; JEDI and action-only receive the same eight human-hand demonstrations, while instruction-only masks the demonstration slots.
The post-training ablations remove keypoint tokens or predictive losses with the backbone, data and 3k-step schedule fixed.
R1 has two conditions with 20 trials in total, and R2 and R3 have one condition each (10 trials).

\begin{table}[!htbp]
\centering\small
\begin{tabular}{@{}lccc@{}}
\toprule
Method & R1 subset & R2 subset & R3 subset \\
\midrule
JEDI, $N{=}8$ & \draftvalue{$15/20$} & \draftvalue{$8/10$} & \draftvalue{$5/10$} \\
Action-only, $N{=}8$ & \draftvalue{$11/20$} & \draftvalue{$5/10$} & \draftvalue{$2/10$} \\
Instruction-only & \draftvalue{$10/20$} & \draftvalue{$4/10$} & \draftvalue{$4/10$} \\
No keypoint tokens & \draftvalue{$11/20$} & \draftvalue{$5/10$} & \draftvalue{$5/10$} \\
No predictive loss & \draftvalue{$9/20$} & \draftvalue{$6/10$} & \draftvalue{$3/10$} \\
\bottomrule
\end{tabular}
\caption{\textbf{Matched real-robot context and post-training controls} (successes/trials; 10 per condition). Subsets use the R1/R2/R3 predicates of Table~\ref{tab:real}.}
\label{tab:real-controls}
\end{table}

\paragraph{Safety.}
The arms operate within software workspace, velocity and torque limits, an operator holds an emergency stop during every rollout, and a rollout is aborted when a limit is reached.

\FloatBarrier
\section{Additional Evaluation}
\label{app:minimal}

\begin{figure}[t]
\centering
\includegraphics[width=\columnwidth]{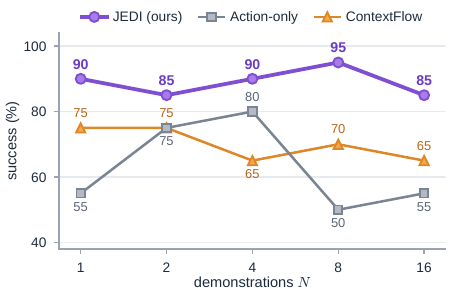}
\caption{\textbf{Matched context-size comparison on LIBERO-Object.} Identical instructions, two held-out tasks and 20 paired trials per point.}
\label{fig:matched-curves}
\end{figure}

\paragraph{Context size and input controls.}
The matched LIBERO-Object comparison uses two tasks from the four-task held-out split and 20 paired trials per condition (two tasks, five support draws and two query layouts).
JEDI, action-only and ContextFlow receive identical instructions and nested support sets at $N\in\{1,2,4,8,16\}$ (Figure~\ref{fig:matched-curves}).
The no-instruction condition removes the text at $N{=}1$; instruction-only ($N{=}0$) masks the demonstration slots while retaining the same text and checkpoint.
Query layouts and success predicates are shared across these controls.

\begin{table}[t]
\centering\small
\begin{tabular}{@{}lcc@{}}
\toprule
Method & No instruction & Instruction only \\
\midrule
JEDI & \draftvalue{65.0} & \draftvalue{25.0} \\
Action-only & \draftvalue{60.0} & \na \\
ContextFlow & \draftvalue{35.0} & \na \\
\bottomrule
\end{tabular}
\caption{\textbf{Input controls on LIBERO-Object} (success, \%; 20 paired trials per cell). Instruction-only retains the JEDI policy with its demonstration context masked. The order-instruction controls are in Table~\ref{tab:language}.}
\label{tab:input-controls}
\end{table}

\paragraph{Uncertainty and support selection.}
At $N{=}1$, JEDI minus ContextFlow is 7.1/7.2/6.5/8.5 points on Spatial/Object/Goal/RLBench (Table~\ref{tab:main}).
The corresponding paired 95\% intervals are \draftvalue{$[2.5,11.7]$}, \draftvalue{$[0.7,13.7]$}, \draftvalue{$[0.9,12.1]$}, \draftvalue{$[3.2,13.8]$}, respectively.
Intervals resample training seeds and paired task/support/layout indices, conditional on the evaluated tasks.
Selection with the three checkpoints is reported in Table~\ref{tab:selection}.

\paragraph{Language semantics.}
Two held-out LIBERO-Order pairs are each tested in both orders, using a canonical instruction and three equivalent forms with \emph{after}, \emph{before} and \emph{not until}.
The 16 instructions each use five query layouts with aligned or conflicting eight-demonstration context.
Success requires endpoint completion and the instructed order; in conflicts the instruction defines the target.
Canonical and paraphrase columns use aligned context (20 and 60 trials); the conflict column pools all forms (80 trials).
The instruction-only reference uses the same instruction/layout cases with its context masked.
Results are in Table~\ref{tab:language} (\S\ref{sec:language}).

\FloatBarrier

\end{document}